\documentclass[letterpaper, 10 pt, conference]{ieeeconf}  %

\IEEEoverridecommandlockouts                              %

\usepackage{cite}
\usepackage{amsmath,amssymb,amsfonts}
\usepackage{algorithmicx}
\usepackage{algorithm}
\usepackage{algpseudocode}
\usepackage{url}
\usepackage{graphicx}
\usepackage{textcomp}
\usepackage{xcolor}
\usepackage{xspace}
\usepackage{subcaption}
\usepackage{multirow}
\usepackage{booktabs}
\usepackage{balance}
\usepackage{eso-pic}
\AddToShipoutPictureBG*{%
  \AtPageLowerLeft{%
    \put(0,30){%
      \makebox[\paperwidth][c]{%
        \parbox[t]{\textwidth}{\footnotesize
          \textcopyright~2026 IEEE. Personal use of this material is permitted.
          Permission from IEEE must be obtained for all other uses, in any current
          or future media, including reprinting/republishing this material for
          advertising or promotional purposes, creating new collective works, for
          resale or redistribution to servers or lists, or reuse of any copyrighted
          component of this work in other works.}}}}}

\newcommand{\fref}[1]{Fig.~\ref{#1}\xspace}

\algrenewcommand{\algorithmiccomment}[1]{\texttt{// #1}}
\newenvironment{itquote}
{\begin{quote}\itshape}
{\end{quote}}
\title{\LARGE \bf
Efficacy of Monocular Depth Estimation in Lunar Environments 
}

\author{Michael W. Lanighan \and Kenji Brameld \and Seth Gee%
\thanks{Authors are with TRACLabs Inc., 1331 Gemini Street; Houston, TX 77058. Corresponding author: {\tt \small mlanighan@traclabs.com}  
. This work was supported by NASA contract 80NSSC25C0381.}}

\begin{document}

\maketitle
\thispagestyle{empty}
\pagestyle{empty}

\begin{abstract}
Recent advances in monocular depth estimation have demonstrated highly accurate 3D reconstructions within in-distribution terrestrial domains. This study investigates the efficacy of several state-of-the-art approaches to reconstruct out-of-distribution lunar environments using the NASA Ames Polar Optical Lunar Analog Reconstruction (POLAR) stereo dataset and Apollo 17 surface imagery. By leveraging these datasets, we establish a robust analog-to-field validation pipeline that evaluates quantitative performance against ground-truth physical analogs and qualitative performance against historical field data. Our results indicate that select state-of-the-art techniques can successfully generalize to produce high-fidelity relative 3D reconstructions of the unstructured lunar surface.
\end{abstract}

\section{Introduction}
Humanity has decades of experience in conducting and overseeing remote space operations, both in terms of human spaceflight and remote robotic assets. Even with this extensive experience, low Earth orbit (LEO) operations on the International Space Station (ISS) for the Mobile Servicing System (MSS) present challenges for operators related to situational awareness. NASA Astronaut Nick Hague recently outlined the state of operations in LEO: 
\begin{itquote}
    Robotics and space exploration go hand in hand. It takes years of training and practice to learn how to digest multiple 2D views at different angles to determine if the path will be clear for the arm to move. We even use the HECA cams on EV1 and EV2 to help.\footnote{\url{https://www.instagram.com/p/DFsc9bjxp-s}}
\end{itquote}
Astronaut Hague's statement emphasizes a key challenge in remote robot operations: the lack of tools to maintain effective, task-relevant situational awareness. The robot operator station on the ISS is dominated by 2D camera views. Currently, operators must ingest these views and create mental models that fuse multiple sources of information together, a skill that, as Astronaut Hague points out, can take years to master. The heavy cognitive load this induces is a common pitfall of remote robot operations. Research has shown that unifying this information in 3D environments enhances task outcomes~\cite{norton2017analysis, pryor2020interactive}. %

Robot operations from mission control at NASA JSC use a similar control station as that on station, but face further challenges brought by downlink/bandwidth constraints. Due to the finite number of downlinks and fixed bandwidth available, not all camera views needed to support a remote operation may be available. While tools to support operations, including 3D views~\cite{dershowitz2002bird} and special purpose capability for docking, have been developed~\cite{lucier2021iss}, integrated 3D solutions to handle general operations are lacking. In upcoming Artemis missions, NASA envisions extended periods of uncrewed activity on the lunar surface. During this time, NASA aims to use robotic assets to perform post-crew-departure activities, carry out scientific and exploration objectives, and relocate themselves and supplies to future landing sites to support crewed missions. Compared to LEO operations, it is anticipated that lunar operations will have decreased  available bandwidth and downlinks, which will only further exacerbate issues surrounding situational awareness for remote operators. New tools are required to support both remote operators and autonomous systems to maintain sufficient situational awareness to execute at the envisioned goal cadences.

\begin{figure}[t]
    \centering
    \begin{minipage}[c]{0.33\columnwidth}
        \centering
        \includegraphics[height=1.0in]{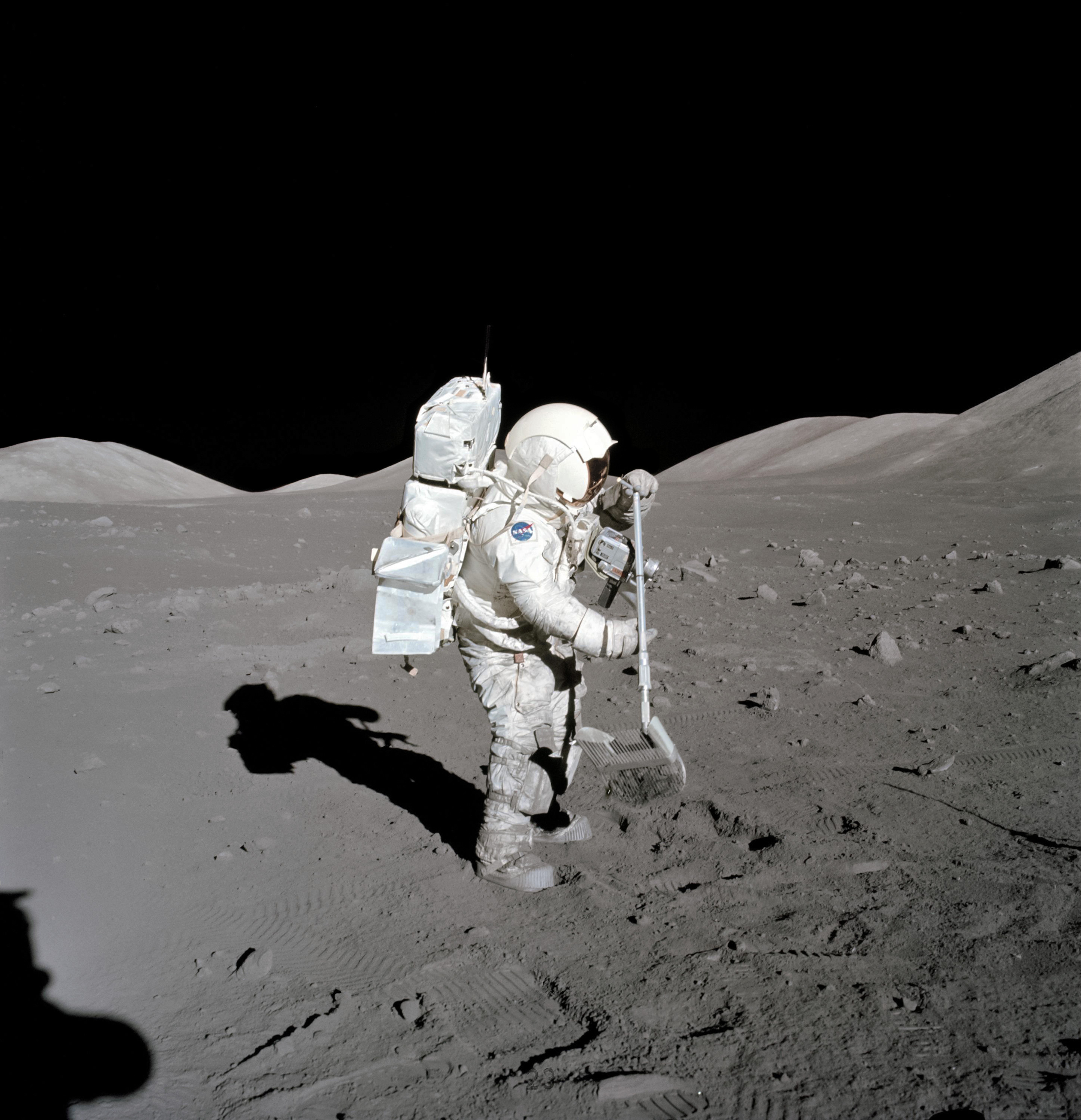}
        \captionof*{figure}{\textbf{Input Image}} %
    \end{minipage}
    \hfill %
    \begin{minipage}[c]{0.65\columnwidth}
        \centering       

        \begin{subfigure}[b]{\linewidth}
            \includegraphics[height=1.0in]{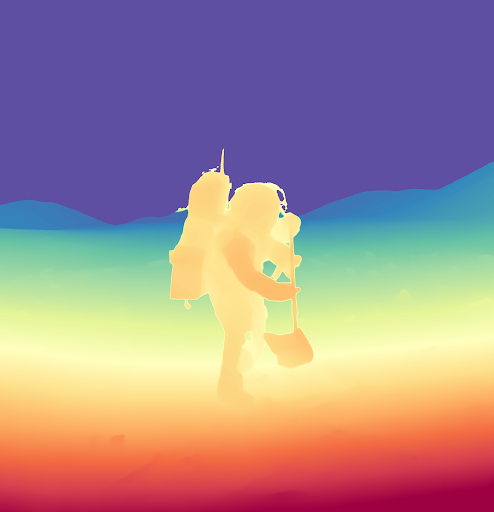} \hfill %
            \includegraphics[height=1.0in]{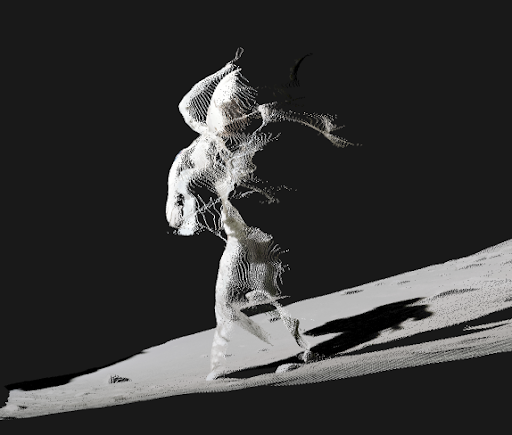} 
            \caption*{\textbf{MoGe-2}} %
        \end{subfigure}   

        \begin{subfigure}[b]{\linewidth}
            \includegraphics[height=1.0in]{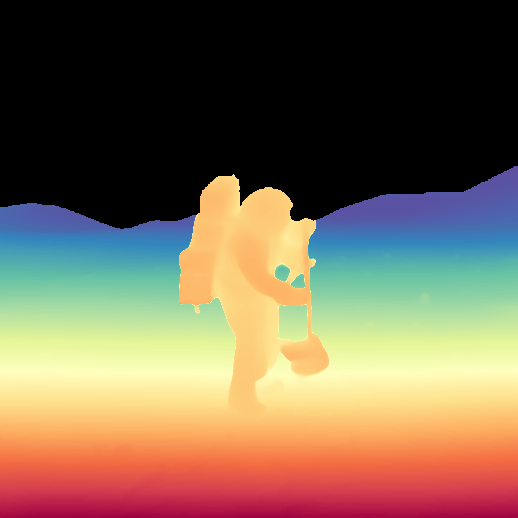} \hfill
            \includegraphics[height=1.0in]{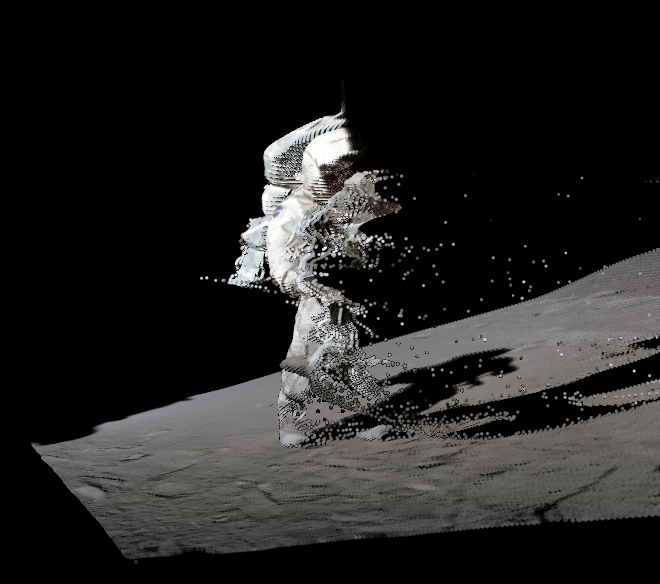} 
            \caption*{\textbf{OmniVGGT}}
        \end{subfigure}
        
        \begin{subfigure}[b]{\linewidth}
            \includegraphics[height=1.0in]{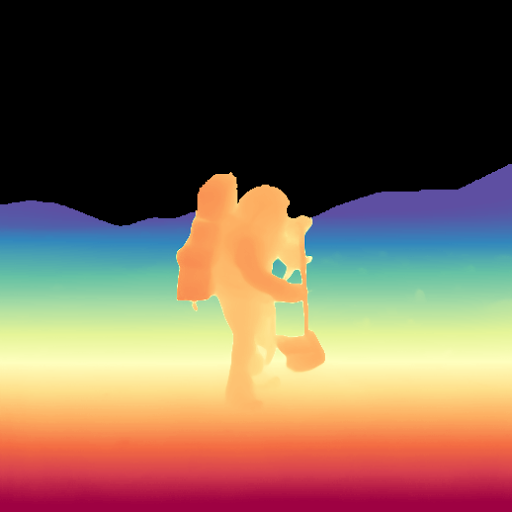} \hfill
            \includegraphics[height=1.0in]{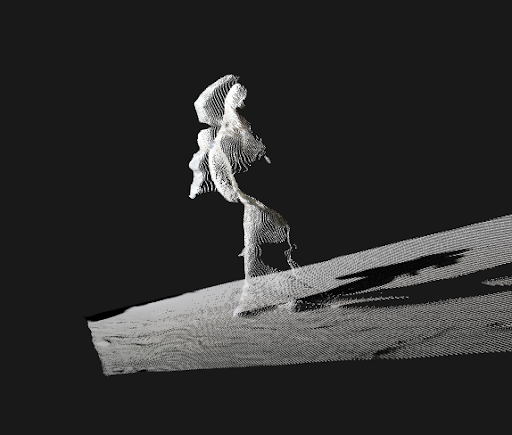} 
            \caption*{\textbf{VGGT}}
        \end{subfigure}

    \end{minipage}

    \caption{Input Apollo 17 image (left) and the resulting depth images and point clouds (middle and right columns) generated by the specified monocular depth estimation techniques. Note the consistent reconstruction of the barren, textureless lunar landscape, even in the presence of shadows.}
    \label{fig:astro}
\end{figure}

A promising emerging area that can help address these issues is monocular depth estimation (MDE). These approaches consume a 2D image and reconstruct a 3D scene for operators to inspect. Moreover, relying on a single camera reduces the need for stereo pairs or time of flight sensors to reconstruct 3D representations of the remote scene, providing the additional benefit of lowered bandwidth requirements. However, state of the art MDE approaches are trained using massive image datasets of terrestrial images. The efficacy of such approaches in domains that are out-of-distribution with these large training datasets, such as the lunar surface, is currently unknown. To address this, this study evaluates several state-of-the-art techniques to determine their efficacy in lunar environments, leveraging images from the NASA Polar Optical Lunar Analog Reconstruction (POLAR) stereo dataset~\cite{wong2017polar} and Apollo 17 mission operations~\cite{jones_apollo17_images} (see \fref{fig:astro}).

\section{Related Work}
While advancements in LiDAR and RGB-D cameras for space systems have been made, it is likely that RGB cameras will be the prevalent form of visual feedback, both on rovers and lunar infrastructure. As such, techniques to create depth estimates and 3D representations using traditional cameras will be needed.

\subsection{Stereogrammetry}
\label{sec:asp}
Recent NASA missions, including the VIPER mission, have targeted onboard stereo pairs to provide 3D information about the operating environment. VIPER CONOPS relies on sending RGB images back to ground control, where they will be used to construct 3D models of the rover's environment via stereogrammetry~\cite{bualat2023intelligent}. These 3D models will then be used by operators when planning rover operations. The core libraries used in this reconstruction were open sourced in the NASA Ames stereo pipeline (ASP)~\cite{beyer2018ames}. Stereo techniques such as those in ASP rely on matching features across multiple images using known offsets between cameras to estimate depth. Due to their fundamental reliance on geometry, these techniques require at least two sensors/images with overlapping visual coverage to obtain depth estimates. Our ASP baseline was explicitly tuned for lunar terrain. Leveraging the intrinsic and extrinsic calibration provided by the POLAR dataset, we configured ASP with the More Global Matching (MGM) algorithm to handle reduced surface texture, alongside sub-pixel refinement and a 9x9 correlation kernel to preserve sharp details on rocks and craters.

\subsection{Monocular Depth Estimation}
\label{sec:monocular}
An alternative approach to estimate depth from RGB images is to use monocular depth estimation techniques. This broad family of techniques trains a neural network to estimate depth in RGB images using large training sets. Several leading techniques have been open-sourced with permissive licenses. These techniques are attractive for remote operations using monocular cameras, as they can produce depth estimates without requiring overlapping visual features from multiple cameras. Provided that similar CONOPS as those proposed in VIPER are followed~\cite{bualat2023intelligent}, cameras could be streamed to remote operators (human or autonomous systems), and depth estimates can be created by feeding incoming images through the monocular depth networks to provide 3D context. Such an approach would cut bandwidth and downlink requirements in half compared to traditional stereogrammetry approaches. While concurrent benchmarking \cite{sekmen2026lumon} evaluates the algorithmic performance of MDE for future autonomous navigation, our work specifically evaluates MDE as a practical solution to overcome these bandwidth limits in remote teleoperation and supervision. Broadly speaking, MDE approaches can be grouped into two classes: \textit{single view reconstruction} and \textit{multi-view reconstruction}. In this study, we evaluate the following approaches:
\subsubsection{Single View Reconstruction} Approaches take as input a single image and estimate depth relative to the imager.
\textbf{Marigold~\cite{ke2024repurposing}:} A diffusion-based (generative) approach that denoises a latent image encoding to produce a depth image.
\textbf{Depth Anything V2~\cite{yang2024depthb}:} Uses Vision Transformer (ViT), DINOv2~\cite{oquab2023dinov2} and DPT~\cite{ranftl2021vision}, architectures to distill a large teacher model trained on synthetic data into a robust estimator using pseudo-labeled real-world imagery. 
\textbf{MoGe-2~\cite{wang2025moge}:} Uses a DINOv2 backbone with multiple decoder heads to directly output affine-invariant relative point maps via a CNN and absolute metric scale via an MLP conditioned on the global CLS token.  
\subsubsection{Multi-View Reconstruction} Approaches consume multiple images to reconstruct a coherent, fused 3D scene. 
\textbf{VGGT~\cite{wang2025vggt}:} Uses a DINOv2 backbone with alternating frame-wise and global self-attention to encode multiple images. It decodes them with multiple heads (a DPT head for points, depth, and tracking, and a camera head for both intrinsics and extrinsics) to directly reconstruct a scene in a single forward pass.
\textbf{MapAnything~\cite{keetha2025mapanything}:} Encodes input images using DINOv2, fusing optional supporting geometric inputs (calibration, poses, depth) into the same latent space. It leverages an alternating attention backbone and features task-specific heads (DPT, pose, and scale) to directly regress factored metric 3D geometry. 
\textbf{OmniVGGT~\cite{peng2025omnivggt}:} Built on a similar structure to VGGT, it introduces a GeoAdapter mechanism with zero-initialized convolutions to safely inject arbitrary auxiliary geometric modalities (depth, camera intrinsics, and extrinsics) without disrupting the pre-trained visual backbone.

\section{Methodology}
To establish the efficacy of the investigated MDE approaches, we leveraged two datasets of representative lunar imagery. The first was a subset of images from the NASA Ames POLAR Stereo dataset~\cite{wong2017polar}. These images are part of a dataset built to measure stereo reconstruction performance in simulated lunar terrains and contain ground truth LiDAR scans of each scene. Scenes are captured from multiple perspectives and lighting conditions. We selected a subset of these images that we felt were representative use cases and would challenge reconstruction ability due to a lack of textures and the presence of lens flaring/glare. The second dataset consisted of a subset of images of the Lunar surface collected during Apollo 17~\cite{jones_apollo17_images}. While ground truth depth does not exist for these images, they can be used to qualitatively measure reconstruction performance and provide analog-to-field validation of the MDE techniques.

\subsection{Quantitative analysis: NASA POLAR Stereo dataset}
The initial evaluations used NASA's POLAR Stereo dataset to conduct a rigorous quantitative analysis of reconstruction performance. Originally developed for Resource Prospector and later for VIPER, the dataset includes stereo images of various laboratory simulated lunar terrains with varying features under numerous lighting conditions from 3 viewing locations. Ground truth data are provided for each scene using a Faro x330 LiDAR, which is registered to the stereo imaging pair. We evaluated images taken from Positions A and C, with the ``sun" 30 degrees from A. For each terrain, capture location C presents lens flaring that we anticipated would cripple the performance of the MDE approaches. Images used in evaluations span strewn rock fields to barren, textureless scenes to evaluate the MDE approaches under the expected diverse conditions on the lunar surface. The image from the left camera of the stereo pair was fed to the MDE approaches, while ASP leveraged both left and right images to reconstruct depth, with the output aligned to the left image. While some of the multi-view approaches support both multiple camera views and camera calibration data, we withheld these inputs in our analysis and focused solely on blind reconstruction capabilities using a single input image.

Quantitative analysis efforts of the targeted MDE approaches are complicated by the fact that not all MDE techniques evaluated output metric depth; many only output relative depth (i.e., normalized) or can incorrectly estimate the global scale. This leads to difficulties in using an absolute error metric, as high absolute errors may be present for a scene reconstructed perfectly but at a different scale. In this study, we were more interested in an approach's ability to faithfully reconstruct the relative structure in scenes rather than correctly estimate the scale, as the reconstruction could be inspected in isolation or the actual scale could be recovered by a downstream process. The MDE community has developed a metric called Scale-Invariant Logarithmic error~\cite{eigen2014depth} to capture an approach's ability to faithfully reconstruct depth without requiring global scale. Let $y_i$ be the predicted depth and $y^*_i$ be the ground truth depth at pixel $i$. The log-difference $d_i$ between the prediction and the ground truth for a given pixel is defined as:
\begin{equation}
    d_i = \log(y_i) - \log(y^*_i)
\end{equation}

Given $n$ total pixels with valid ground truth measurements, the original Scale-Invariant Logarithmic (SILog) error is calculated as the variance of the log-differences:
\begin{equation}
    \text{SILog}_{original} = \frac{1}{n} \sum_{i=1}^{n} d_i^2 - \frac{1}{n^2} \left( \sum_{i=1}^{n} d_i \right)^2
\end{equation}

This measure captures relative errors in the reconstruction. Specifically, the mean log-differences $\alpha =\frac{1}{n} \sum_{i=1}^{n}d_i$ capture the scale $e^{-\alpha}$ that best aligns any prediction with ground truth. The variance, therefore, captures the remaining reconstruction errors.  However, $\text{SILog}_{original}$ is a variance; its units are squared logarithmic differences, which lack intuitive physical meaning. In standard benchmarks~\cite{uhrig2017sparsity}, a scaled variant (which we will call SILog to match the research community) is used. This variant applies a square root to convert the variance into a linear standard deviation and multiplies it by a constant factor of 100:
\begin{equation}
    \text{SILog} = 100 \times \sqrt{ \frac{1}{n} \sum_{i=1}^{n} d_i^2 - \frac{1}{n^2} \left( \sum_{i=1}^{n} d_i \right)^2 }
\end{equation}

The resulting $\text{SILog}$ value encodes how tight the distribution of structural errors in the reconstruction is in log space. Specifically, converting back to linear space via $e^{\text{SILog}/100}$ provides the $1\sigma$ multiplicative error factor $\rho$, indicating that approximately 68\% of the scale-aligned pixel depth reconstructions ($y_{aligned} = y \cdot e^{-\alpha}$)  fall within ${y^*}/{\rho} \le y_{aligned} \le \rho \cdot y^* $ relative to the ground truth $y^*$. The $2\sigma$ bounds extend to $\rho^2$, as outlined in Table \ref{tab:silog_grounding}.

\begin{table}[htbp]
\centering
\caption{Statistical Interpretation of $\text{SILog}$ values}
\label{tab:silog_grounding}

\begin{tabular}{rcc}
\toprule
\textbf{$\text{SILog}$} & \textbf{$1\sigma$ Error Factor ($\rho$)} & \textbf{$2\sigma$ Error Factor ($\rho^2$)} \\\midrule
2.5 & $1.03\times$   & $1.05\times$ \\
5   & $1.05\times$   & $1.11\times$ \\
10  & $1.11\times$   & $1.22\times$ \\
50  & $1.65\times$   & $2.72\times$ \\
100 & $2.72\times$   & $7.39\times$ \\
\bottomrule
\end{tabular}
\end{table}

\subsection{Qualitative analysis: Apollo 17 Images}
The second evaluation uses images of the Lunar Surface collected during Apollo 17 to qualitatively compare the best performing MDE approaches. Qualitative analysis inspected the depth reconstruction fidelity in terms of overall structure and relative noise to validate our findings using the NASA POLAR stereo dataset. We selected two images from the dataset: one that prominently features Astronaut Schmitt conducting sample collection activities against a barren lunar terrain and a second that prominently features the texture rich rear of the lunar roving vehicle (LRV) after the ``fender repair'' incident, with Schmitt in the LRV again backdropped by barren lunar terrain. It is expected that these images will challenge MDE approaches, as they are unlikely to be representative of the training datasets used to create each model.

\section{Results}
\begin{figure}[!ht]
    \centering
    \includegraphics[width=\linewidth]{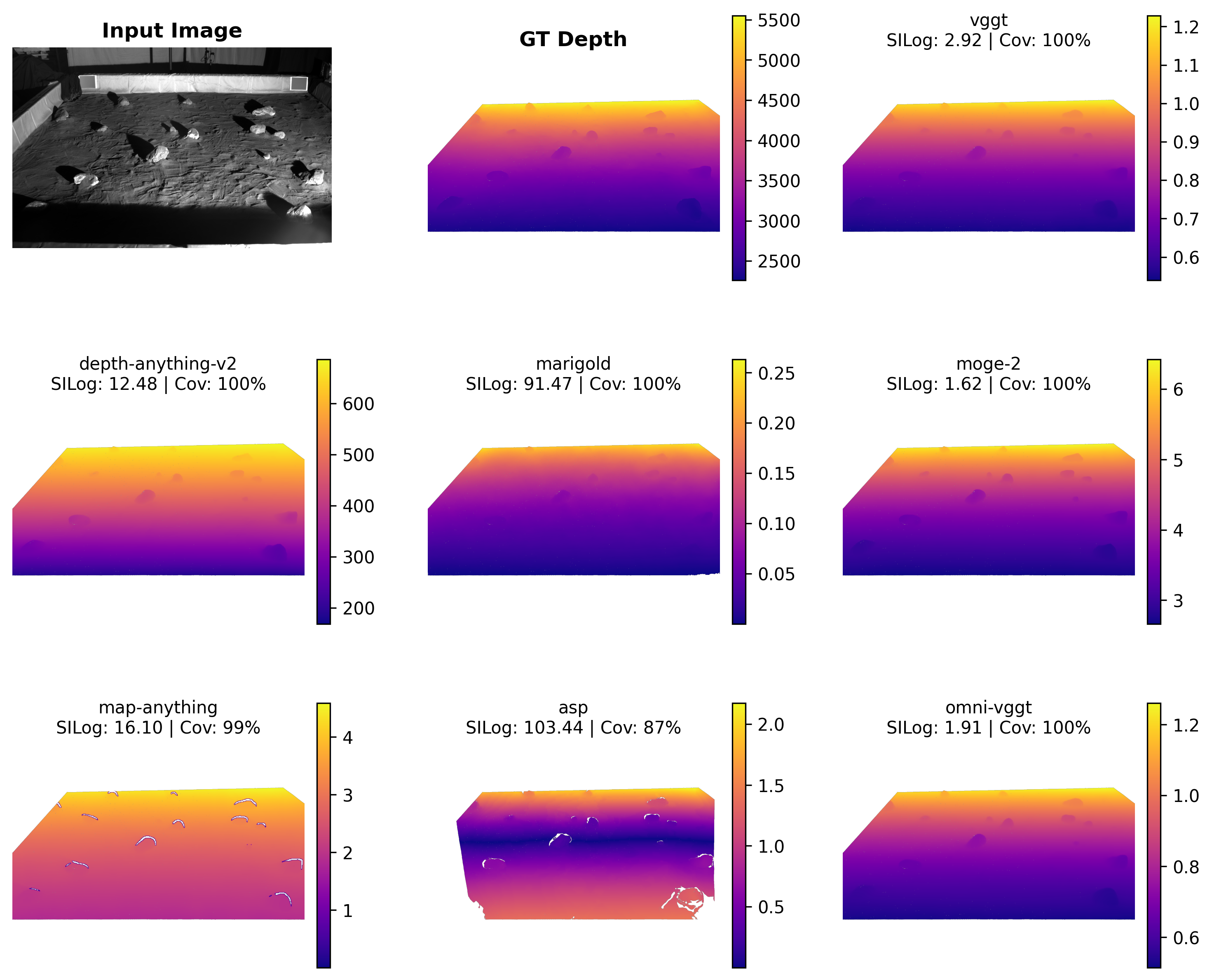}
    \caption{Reconstructions for Terrain 1, capture position A of the POLAR stereo dataset featuring a rock-strewn terrain. The figure shows the input, the masked ground truth depth, and the masked reconstructions of each approach used to calculate SILog.}
    \label{fig:t1a}
\end{figure}

\begin{figure}[!ht]
    \centering
    \includegraphics[width=\linewidth]{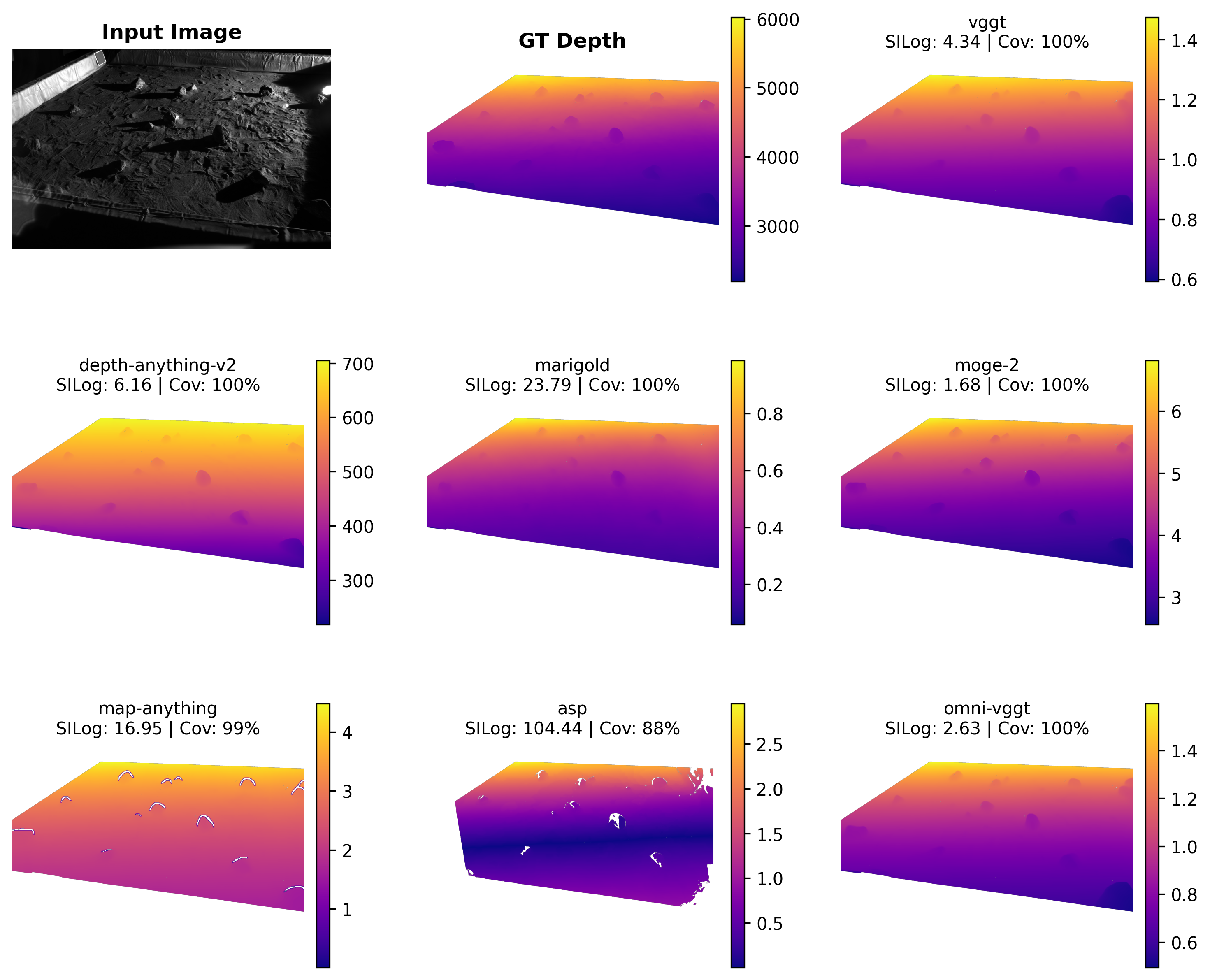}
    \caption{Reconstructions for Terrain 1, capture position C of the POLAR stereo dataset featuring a rock-strewn terrain with lens flaring. The figure shows the input, the masked ground truth depth, and the masked reconstructions of each approach used to calculate SILog.}
    \label{fig:t1c}
\end{figure}

\begin{figure}[!ht]
    \centering
    \includegraphics[width=\linewidth]{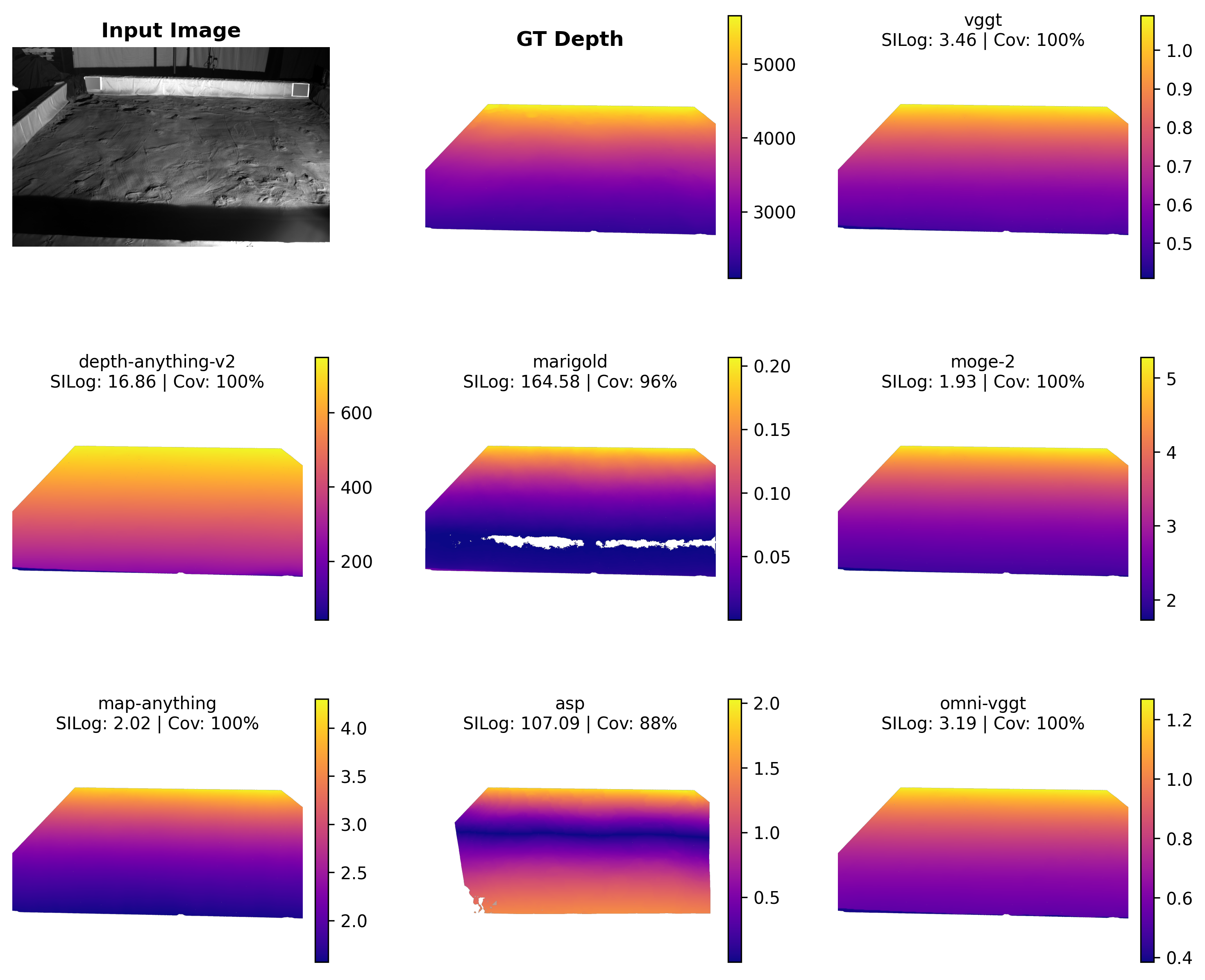}
    \caption{Reconstructions for Terrain 6, capture position A of the POLAR stereo dataset featuring a barren terrain. The figure shows the input, the masked ground truth depth, and the masked reconstructions of each approach used to calculate SILog.}
    \label{fig:t6a}
\end{figure}

\begin{figure}[!ht]
    \centering
    \includegraphics[width=\linewidth]{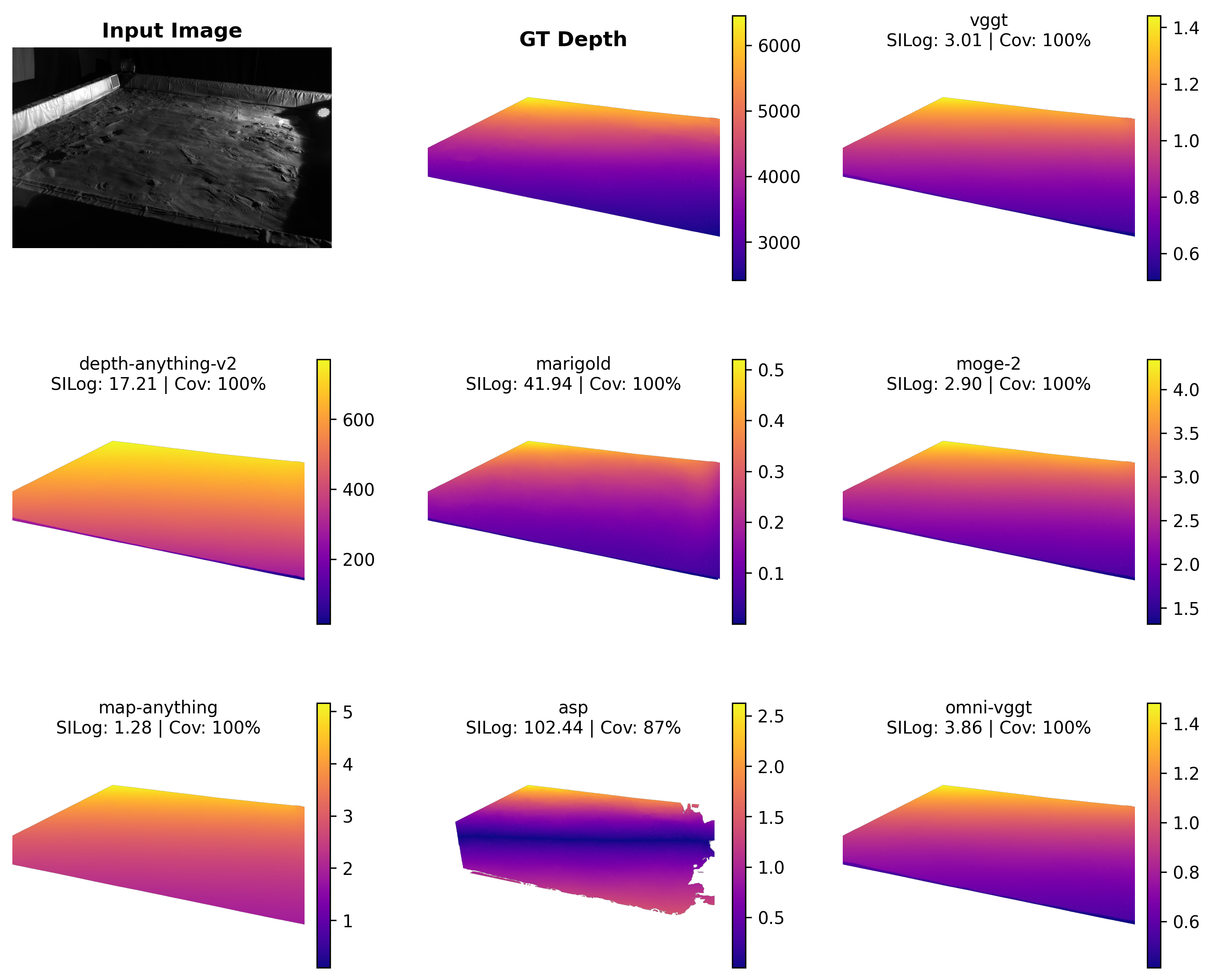}
    \caption{Reconstructions for Terrain 6, capture position C of the POLAR stereo dataset featuring a barren terrain with lens flaring. The figure shows the input, the masked ground truth depth, and the masked reconstructions of each approach used to calculate SILog.}
    \label{fig:t6c}
\end{figure}

\begin{figure}[!ht]
    \centering
    \includegraphics[width=\linewidth]{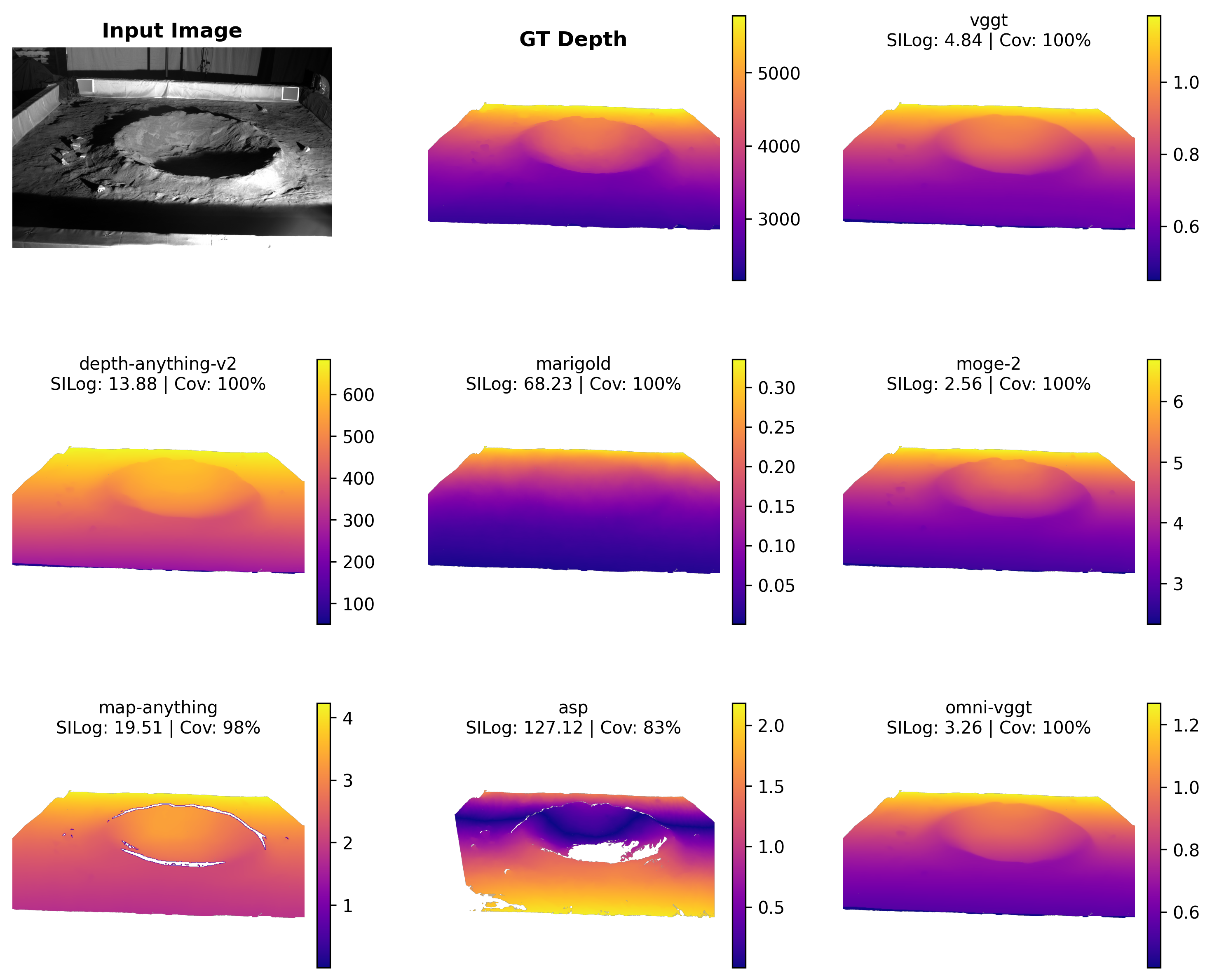}
    \caption{Reconstructions for Terrain 11, capture position A of the POLAR stereo dataset featuring a large crater. The figure shows the input, the masked ground truth depth, and the masked reconstructions of each approach used to calculate SILog.}
    \label{fig:t11a}
\end{figure}

\begin{figure}[!ht]
    \centering
    \includegraphics[width=\linewidth]{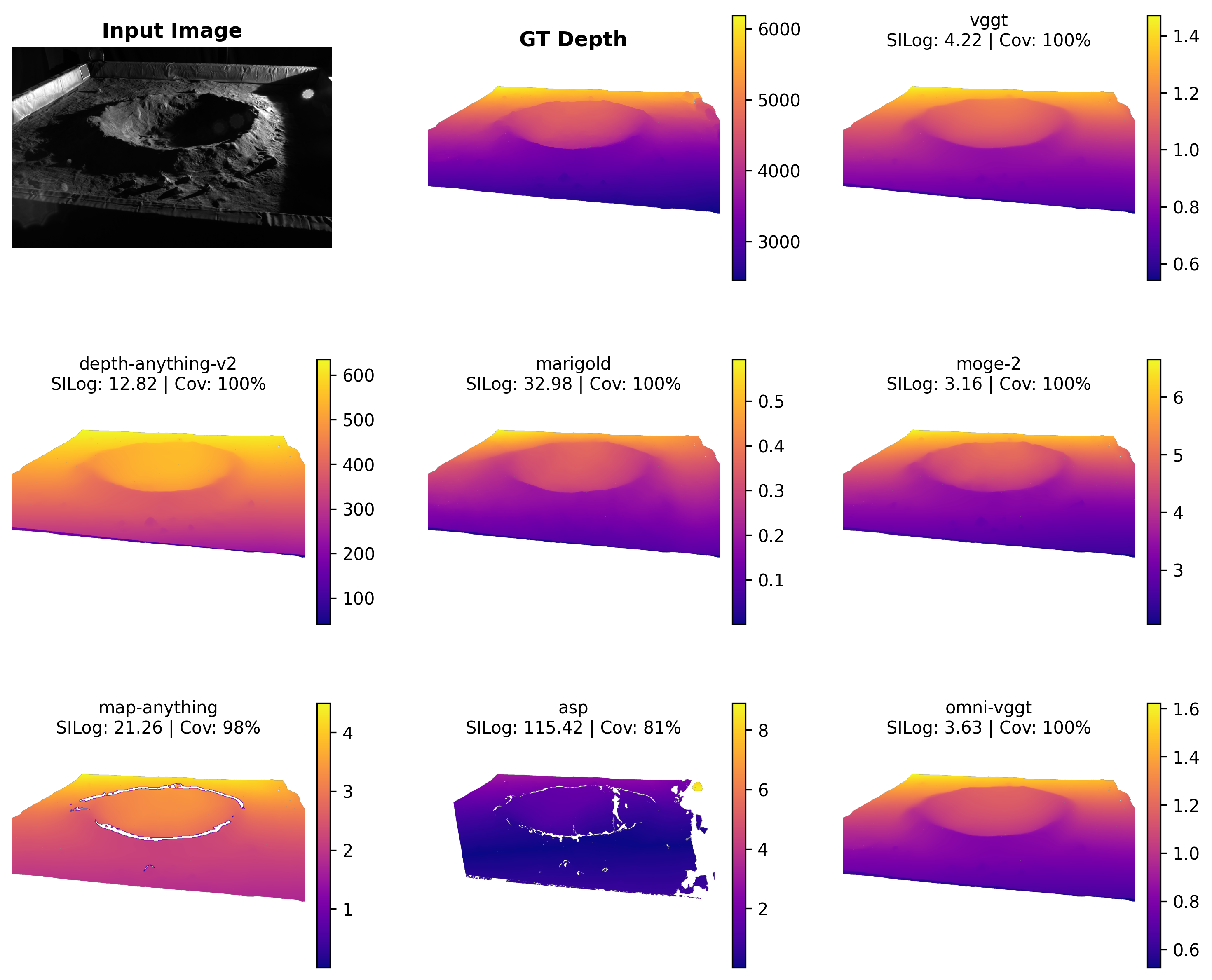}
    \caption{Reconstructions for Terrain 11, capture position C of the POLAR stereo dataset  featuring a large crater with lens flaring. The figure shows the input, the masked ground truth depth, and the masked reconstructions of each approach used to calculate SILog.}
    \label{fig:t11c}
\end{figure}

\fref{fig:t1a} - \fref{fig:t11c} highlight reconstruction results for Terrains 1, 6, and 11 from the POLAR stereo dataset. Each figure shows the input, the masked ground truth depth, and the masked reconstructions of each approach used to calculate coverage and SILog values.  Table \ref{tab:results_detail} highlights the SILog values for each approach and test image, while Table \ref{tab:results_summary} provides summary measures. As can be seen in \fref{fig:t1a} to \fref{fig:t11c}, not all methods estimate depth for each pixel. For each approach, only pixels with depth estimates were used to calculate SILog. %

\begin{table}[htbp]
\centering
\caption{Detailed SILog Results per Terrain and Capture Position}
\label{tab:results_detail}
 \resizebox{\columnwidth}{!}{ 
\begin{tabular}{lrrrrrr}
\toprule
 & \multicolumn{2}{c}{T01-Rocks} & \multicolumn{2}{c}{T06-Smooth} & \multicolumn{2}{c}{T11-FreshCrater} \\
\cmidrule(lr){2-3} \cmidrule(lr){4-5} \cmidrule(lr){6-7}
Approach & Pos A & Pos C & Pos A & Pos C & Pos A & Pos C \\
\midrule
MoGe-2             & 1.62   & 1.68   & 1.93   & 2.90   & 2.56   & 3.16   \\
OmniVGGT           & 1.91   & 2.63   & 3.19   & 3.86   & 3.26   & 3.63   \\
VGGT               & 2.92   & 4.34   & 3.46   & 3.01   & 4.84   & 4.22   \\
MapAnything        & 16.10  & 16.95  & 2.02   & 1.28   & 19.51  & 21.26  \\
Depth Anything v2  & 12.48  & 6.16   & 16.86  & 17.21  & 13.88  & 12.82  \\
Marigold           & 91.47  & 23.79  & 164.58 & 41.94  & 68.23  & 32.98  \\
ASP                & 103.44 & 104.44 & 107.09 & 102.44 & 127.12 & 115.42 \\
\bottomrule
\end{tabular}
}
\end{table}

\begin{table}[htbp]
\centering
\caption{Overall Performance Metrics by Approach}
\label{tab:results_summary}
\resizebox{\columnwidth}{!}{ 
\begin{tabular}{lrrr}
\toprule
Approach & Mean SILog $\downarrow$ & Std Dev & Mean Coverage (\%) $\uparrow$ \\
\midrule
MoGe-2              & 2.31   & 0.60  & 100.00 \\
OmniVGGT            & 3.08   & 0.65  & 100.00 \\
VGGT                & 3.80   & 0.71  & 100.00  \\
MapAnything         & 12.85  & 8.10  & 99.12  \\
Depth Anything v2   & 13.24  & 3.65  & 100.00 \\
Marigold            & 70.50  & 47.79 & 99.27  \\
ASP                 & 109.99 & 8.77  & 85.60 \\
\bottomrule
\end{tabular}
}
\end{table}

Our results indicate that all ViT-based approaches perform quite well in reconstructing out-of-distribution data of lunar terrain, with the best performing approach (MoGe-2, mean SILog value of 2.31) resulting in a $2\sigma$ error factor of $1.047$ (indicating $\approx95\%$ of aligned pixel depths $y_{aligned}$ fall within $y^*/1.047 \leq y_{aligned} \leq  1.047y^{*}$ of ground truth $y^*$). This result demonstrates that while the reconstruction scale may be incorrect, the overall structure is largely consistent with physical reality. The diffusion-based approach (Marigold) did not perform as well on the lunar data, indicating that the lunar environment diverged too far from its training data to reliably estimate depths. This is likely because diffusion-based approaches rely heavily on generative priors tied to the latent distribution of their terrestrial training data, making them prone to failure when presented with strictly out-of-distribution imagery. In contrast, the ViT-based approaches leverage robust, generalizable geometric features extracted by their self-supervised backbones, allowing them to better interpret the novel lunar environment. The stereogrammetry approach ASP performed the worst in our evaluations. Despite lunar-specific tuning, traditional stereogrammetry's reliance on photometric consistency caused widespread matching failures on textureless regolith and severe lens flares. These extreme outliers directly drove the high SILog error variance observed. Of particular note is \fref{fig:t1c}, which featured noticeable lens flaring. In this example, the MDE network architectures were able to successfully estimate depth, while ASP was not. This could be due to the MDE networks compensating for visual artifacts, while ASP relies on the raw features present in the image. 

We evaluated the top 3 methods from the quantitative analysis, MoGe-2, Omni-VGGT, and VGGT, on images from the lunar surface taken during Apollo 17. While these results can only be critiqued qualitatively, as no ground truth depth data exist, this provides analog-to-field validation that these techniques can create effective reconstructions of actual Lunar surface imagery. 

The input image and reconstructions of Astronaut Schmitt on the lunar surface are shown in \fref{fig:astro}. As shown by the depth images, all 3 methods correctly place Schmitt at a depth consistent with his foot locations in the image mid-ground. MoGe-2 correctly separates Schmitt and the background, particularly between his left hand and his helmet visor, his left hand and the chest mounted camera, and the back of his helmet and the suit backpack. The other two approaches fail to correctly separate these items and instead estimate the background patches to be the same distance from the camera as Schmitt. MoGe-2 does appear to generate higher noise than the other approaches, most noticeable around the helmet and backpack. Visualizing the point cloud representations, all three techniques capture the scene, with MoGe-2 providing a higher quality reconstruction of the rocks in the foreground than OmniVGGT or VGGT. OmniVGGT appears to provide the most consistent reconstruction of Schmitt's right-side and captures detail in the helmet at higher fidelity than MoGe-2 or VGGT.

\begin{figure}
    \centering
    \begin{minipage}[c]{0.5\columnwidth}
        \centering
        \includegraphics[height=1.1in]{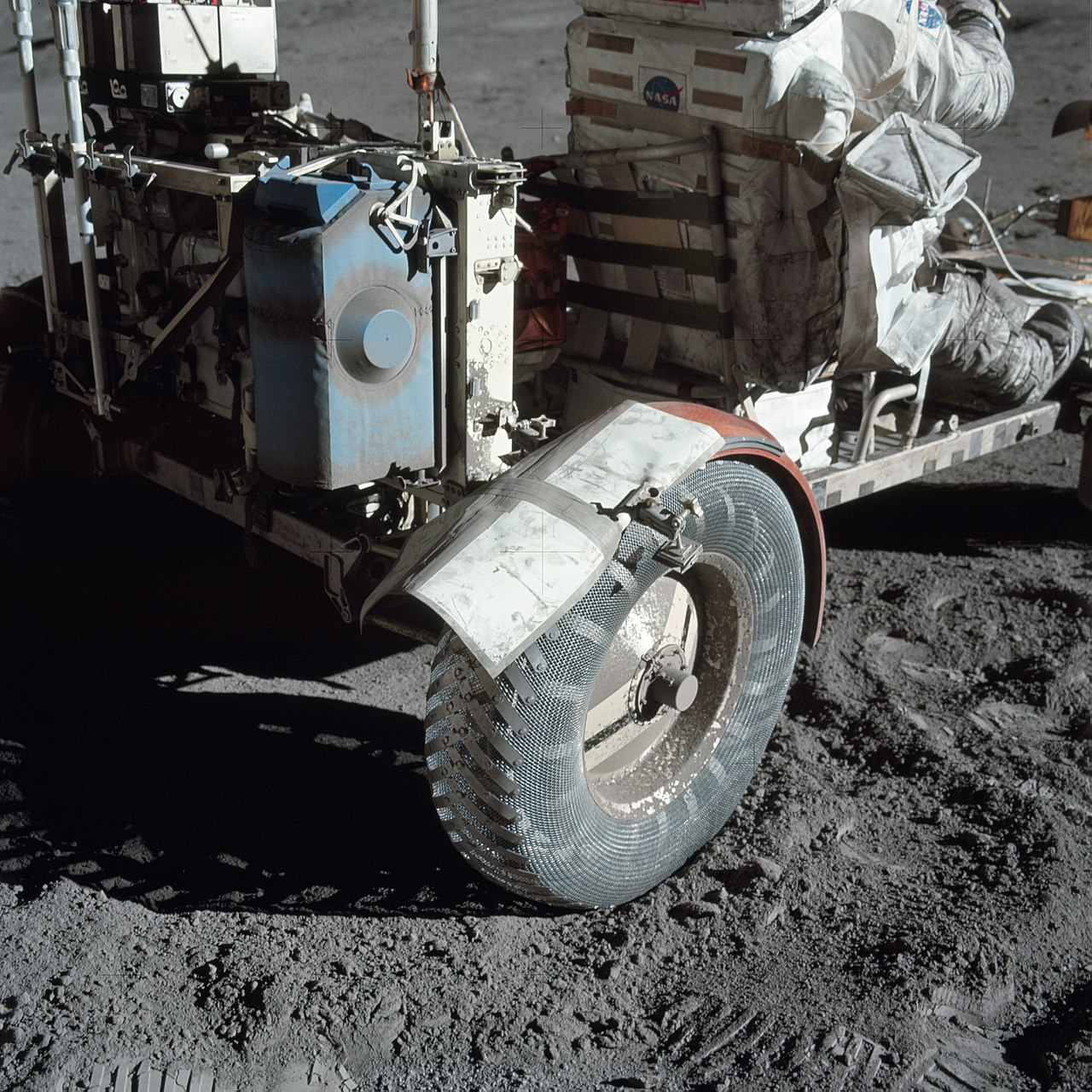}
        \captionof*{figure}{\textbf{Input Image}} %
    \end{minipage}
    \hfill %
    \begin{minipage}[c]{\columnwidth}
        \centering       

        \begin{subfigure}[b]{\linewidth}
            \centering
            \includegraphics[height=1.1in]{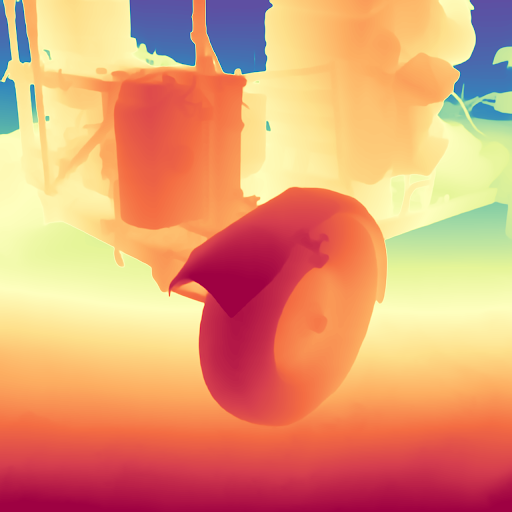} \quad
            \includegraphics[height=1.1in]{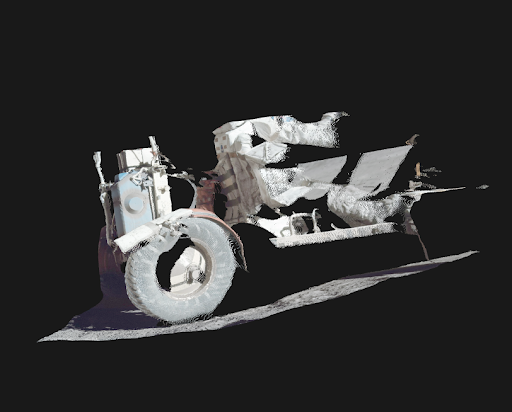} 
            \caption*{\textbf{MoGe-2}}
        \end{subfigure}

        \begin{subfigure}[b]{\linewidth}
            \centering
            \includegraphics[height=1.1in]{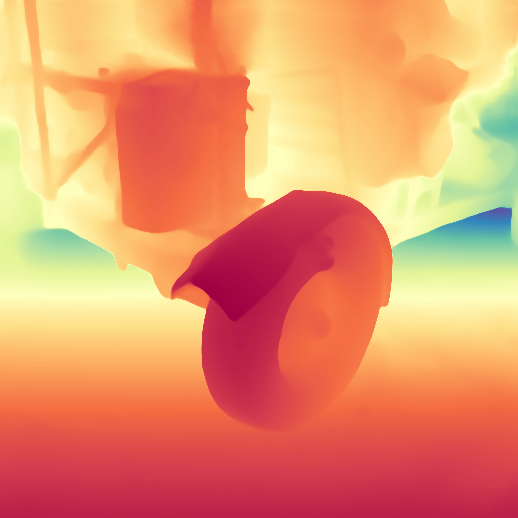} \quad
            \includegraphics[height=1.1in]{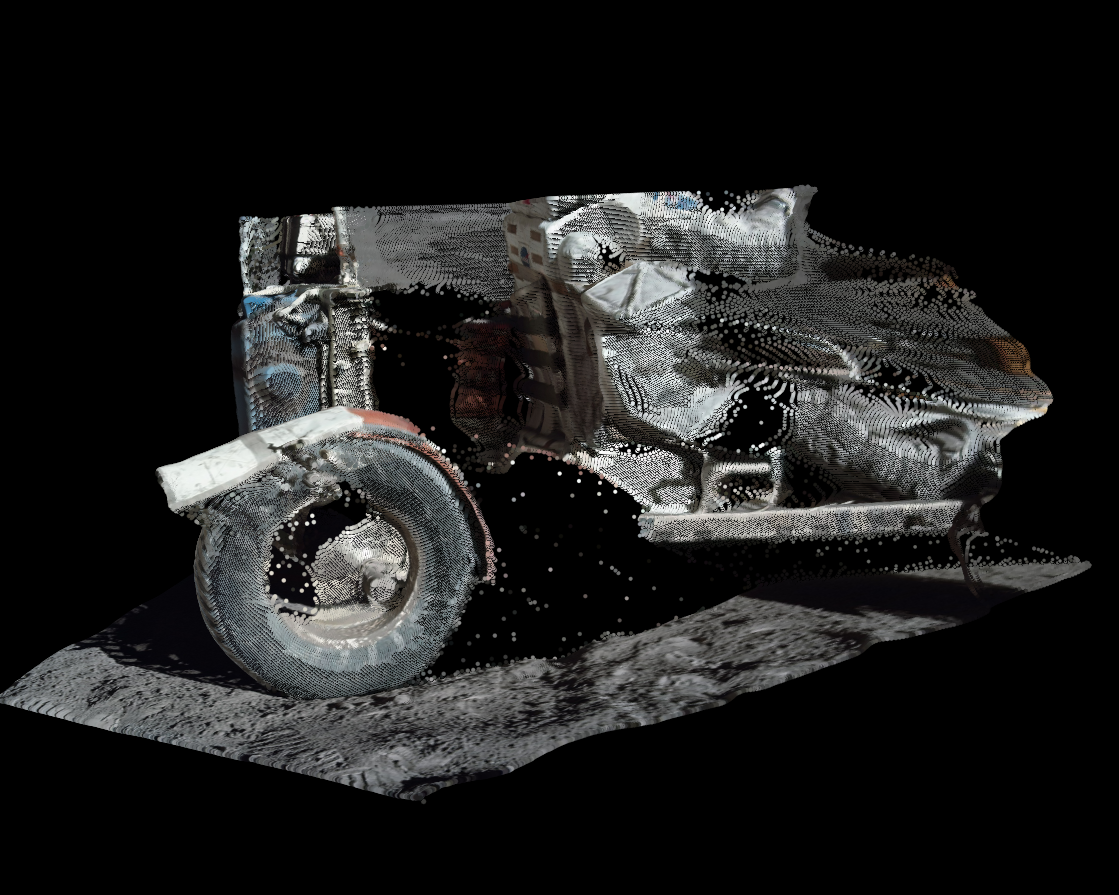} 
            \caption*{\textbf{OmniVGGT}}
        \end{subfigure}
        
        \begin{subfigure}[b]{\linewidth}
            \centering
            \includegraphics[height=1.1in]{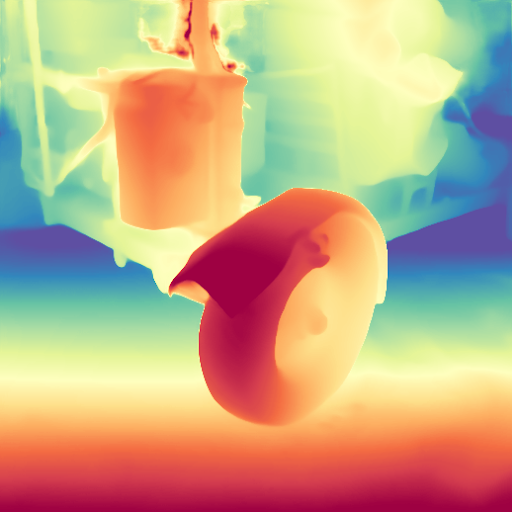} \quad
            \includegraphics[height=1.1in]{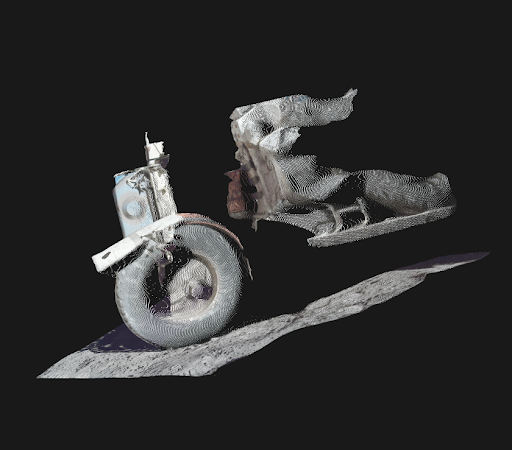} 
            \caption*{\textbf{VGGT}}
        \end{subfigure}

    \end{minipage}

    \caption{Input Apollo 17 image (top) and the resulting depth images (left) and point clouds (right) generated by the specified monocular depth estimation techniques.}
    \label{fig:fender}
\end{figure}

The input and reconstructions of the fender repair incident are shown in \fref{fig:fender}. Of the three techniques, MoGe-2 again correctly separates the LRV from the background, while both OmniVGGT and VGGT more readily blend the background and foreground. This is most noticeable in the OmniVGGT results, which pull the gray lunar background patches between Schmitt and the back of the rover towards the foreground. VGGT does this as well, but less dramatically. No technique quite captures the complete structure perfectly, as evidenced by the distortions in each reconstruction point cloud around the rear right wheel ``hubcap." The reconstructions by MoGe-2 are most consistent with the relative geometries of the physical LRV, which has the seat just ahead of the rear wheel radius. Both OmniVGGT and VGGT appear to overestimate this distance, placing the seat further from the rear wheel than it physically is. These critiques aside, the resulting 3D representations are clearly interpretable given the input image, with the highest fidelity reconstructions generated by MoGe-2.

\section{Conclusions}
Our results indicate that ViT-based MDE approaches such as MoGe-2, OmniVGGT, and VGGT provide high-fidelity relative 3D reconstructions in out-of-distribution lunar environments. A surprising result of our investigation was the relative robustness of MDE approaches  to the impact of lens flaring and shadows. Extreme lighting conditions resulting in optical artifacts and areas of deep shadow are expected on the lunar surface. In the POLAR stereo dataset images, the investigated MDE approaches successfully mitigated lens artifacts that traditional stereogrammetry struggled with. Furthermore, MDE approaches consistently estimated the depth of shadowed areas in both the POLAR stereo and Apollo surface images. This finding provides strong support for leveraging MDE approaches in lunar settings where remote operator introspection is needed before committing to actions. Moreover, obtaining such 3D context from a single image cuts bandwidth needs in half compared to traditional stereogrammetry approaches, which require two views. 

However, the investigated MDE approaches did not consistently output an accurate metric scale, if they output one at all. As such, while the reconstructions were internally consistent, they cannot be readily integrated into additional visualization tools without further downstream processing. For envisioned lunar surface use, particularly in remote teleoperation contexts, this processing could leverage the known dimensions of rover components (e.g., manipulator arm, grippers, etc.) or other infrastructure (e.g., landers or habitats) in the camera's view to dynamically anchor the global scale. Furthermore, even relative reconstructions can be used in isolation to provide vital 3D context, allowing operators to evaluate a scene before committing the additional bandwidth required to downlink stereo pairs for accurate metric depth. The top-performing MDE techniques estimated depth for every pixel in the image. Given this complete coverage, it may also be possible to fuse MDE and stereogrammetry results to fill in any holes from traditional stereo reconstructions that occur due to insufficient feature matching across the pair~\cite{liu2026interplanetary}. 

This study focused on single-view reconstruction, but several of the investigated techniques support multi-view reconstruction. This is a promising direction for future analysis, particularly if these approaches can render internally consistent fused barren lunar landscapes. Early qualitative exploration indicates that multi-view approaches can render such landscapes in the POLAR stereo dataset, but new lunar terrain datasets are needed to fully quantify the errors in the resulting fused scenes. If viable, such multi-view approaches would be capable of providing rich, 3D reconstructions from viewing geometries selected to support remote robot operations~\cite{lanighan2025computing}. 

\bibliographystyle{IEEEtran}
\balance
\bibliography{mybibfile}
\end{document}